\documentclass[review]{elsarticle}
\usepackage{amsmath}
\usepackage{setspace}
\usepackage{ulem}
\usepackage{lineno,hyperref}
\usepackage[T1]{fontenc}
\usepackage[perpage]{footmisc}
\usepackage{booktabs}
\usepackage{subcaption}
\usepackage{algorithm,tabularx}
\usepackage[noend]{algpseudocode}
\usepackage{tikz}
\usepackage{color}
\modulolinenumbers[5]
\newcounter{auxFootnote}
\newlength\tbspace
\newcolumntype{L}{l<{\hspace{\tbspace}}}

\makeatletter
\newcommand{\multiline}[1]{%
  \begin{tabularx}{\dimexpr\linewidth-\ALG@thistlm}[t]{@{}X@{}}
    #1
  \end{tabularx}
}
\makeatother
\algnewcommand{\do}{\textbf{do }}
\algnewcommand\Input{\item[\textbf{Input:}]}%
\algnewcommand\Output{\item[\textbf{Output:}]}%

\journal{Journal of Image and Vision Computing}
\PassOptionsToPackage{draft}{graphicx}

\begin{document}

\begin{frontmatter}

\title{Visual Object Tracking based on Adaptive Siamese and Motion Estimation Network}

\author{Hossein Kashiani\corref{mycorrespondingauthor}}
\cortext[mycorrespondingauthor]{Corresponding author}
\ead{hossein\_kashiyani@alumni.iust.ac.ir}
\author{Shahriar B. Shokouhi}
\ead{bshokouhi@iust.ac.ir}
\address{School of Electrical Engineering, Iran University of Science and Technology, Tehran, Iran}

\begin{abstract}
Recently, convolutional neural network (CNN) has attracted much attention in different areas of computer vision, due to its powerful abstract feature representation. Visual object tracking is one of the interesting and important areas in computer vision that achieves remarkable improvements in recent years. In this work, we aim to improve both the motion and observation models in visual object tracking by leveraging representation power of CNNs. To this end, a motion estimation network (named MEN) is utilized to seek the most likely locations of the target and prepare a further clue in addition to the previous target position. Hence the motion estimation would be enhanced by generating a small number of candidates near two plausible positions. The generated candidates are then fed into a trained Siamese network to detect the most probable candidate. Each candidate is compared to an adaptable buffer, which is updated under a predefined condition. To take into account the target appearance changes, a weighting CNN (called WCNN) adaptively assigns weights to the final similarity scores of the Siamese network using sequence-specific information.
Evaluation results on well-known benchmark datasets (OTB100, OTB50 and OTB2013) prove that
the proposed tracker outperforms the state-of-the-art competitors.
\end{abstract}


\end{frontmatter}

\section{Introduction}
Visual object tracking is the process of determining the target location in a video sequence. It is one of the many remarkable research topics in computer vision that has gained considerable attention over the past decade. It can be applied to numerous applications, including video indexing, activity recognition, augmented reality, vehicle navigation and so forth. Noticeable efforts have been made to improve object tracking accuracy in the past decade; however, it is still an open and active research area since different variations have been imposed on the object of interest during the tracking process.\par
Recently, the success of convolution neural networks (CNNs) has drawn much attention to various areas in computer vision, like image recognition \cite{he2016deep}, object detection \cite{girshick2014rich,lin2017feature} and object tracking \cite{leal2016learning,song2017crest,valmadre2017end,bertinetto2016fully}. CNNs can exploit a wider range of features from an arbitrary target in contrast to low-level hand-crafted features. Low-level convolutional features in earlier CNN layers contain higher spatial resolution appropriate for accurate object localization. While high-level convolutional features in later layers retain a more semantic context \cite{ma2015hierarchical} helpful for distinguishing the foreground object from the background.\par
Generally, a tracking system consists of two fundamental components: 1) An observation model, which models the appearance of the target and validates candidates over time; 2) A motion model, which generates the number of candidates over time \cite{perez2002color,ross2008incremental}.\par
In this paper, we aim to enhance the performance of both these models simultaneously. Based on the observation model, CNN-based visual object tracking is approximately divided into discriminative and generative approaches \cite{li2018deep}. Discriminative methods consider object tracking as binary classification and differentiate the target from the background candidates using either a new CNN model or a pretrained CNN model \cite{he2016deep,DBLP:journals/corr/SimonyanZ14a}. In order to consider the object appearance variations, these methods can be integrated into an online fine-tuning strategy \cite{nam2016learning,wang2016stct}. In the discriminative trackers, online fine-tuning during tracking process can enhance the overall performance of the tracker at the expense of increased computational cost. On the other hand, generative methods seek to learn matching function to make a comparison between the target template and calculated candidates using CNN while maintaining high-speed performance \cite{valmadre2017end,bertinetto2016fully,tao2016siamese}.\par
Siamese network is one of the prominent models in generative methods which achieves promising accuracy as well as high speed. A Y-shaped Siamese network aims to learn the similarity rate between the initial target and the sampled candidates while remaining unchanged over time \cite{bertinetto2016fully,tao2016siamese}. Exploiting one fixed matching function impedes the Siamese-based tracker to take sequence-specific information into account, whereas discriminative trackers take this information into consideration \cite{wang2016stct}. Thus, utilizing nonadjustable matching function and bypassing online updating would make Siamese-based trackers less adaptable in dealing with target appearance changes, background distractions and coexisting of confusing objects. To address these issues, sequence-specific information can be integrated into the Siamese-based model trained offline. In this paper, to reach this objective, a particular Siamese network is followed by a weighting CNN (called \textit{WCNN}) to leverage sequence-specific information. Figure \ref{fig.Wnet} represents the proposed framework and illustrates the collaboration of the WCNN and the Siamese network. The WCNN is updated over time to consider target appearance variations. The influence of sequence-specific information is exerted over the final similarity score of the Siamese network using a weighting mechanism. Integrating sequence-specific information with the Siamese network is worthwhile not only to discern the most similar candidates to target but also to exploit sequence-specific cues in weighting the candidates.\par
In spite of the substantial amount of researches in CNN-based trackers, there are limited numbers of studies in augmenting motion model \cite{sun2016occlusion,yang2017deep}. A naive motion model, which is conventionally employed in tracking algorithms \cite{nam2016learning,guo2017deep,han2017branchout,zhang2016robust}, presumes that the state transition of the target can be modeled by a Gaussian distribution. Consequently, new candidates are generated centered at the former target state by employing a Gaussian distribution. To cope with fast and complex motion, the trackers using this motion model tend to drift away from the target. Furthermore, if the previous estimated location entails background distractions, new candidates are generated based on the previous incorrect target state, and hence the tracker will deviate.\par To address the aforementioned issues, motivated by \cite{nam2016learning,yang2017deep,wang2015visual,hong2015online}, a motion estimation network (named MEN) is utilized to prepare another evidence of target location in addition to the output of the Siamese network. Therefore, motion estimation would be improved by generating a small number of candidates, but near the two most likely locations (outputs of the Siamese and the MEN). In order to take advantage of sequence-specific cues, the last two convolutional layers of the MEN are updated over time. To sum up, the proposed tracker represents a step forward in making the tracker robust against a wide range of challenging scenarios.\par
Evaluations on benchmark datasets (OTB100 \cite{wu2015object}, OTB50 and OTB2013\cite{wu2013online}) demonstrate the superiority of the proposed tracker over the state-of-the-art competitors.

\section{Related Work}
In this section, two main categories of CNN-based trackers related to our method are presented.
\subsection{Discriminative CNN-based Trackers}
With the advent of CNN and its notable success, researchers have tended to take advantage of CNN power in visual object tracking. In \cite{wang2015transferring}, the object of interest is discriminated from the background with a CNN pretrained on a classification dataset. This CNN is then fine-tuned to be adapted to target appearance variations using stochastic gradient descent (SGD). Although Wang et al. \cite{wang2015transferring} obtain acceptable results, using a CNN pretrained on the image classification dataset would not lead to the best performance in visual tracking. Unlike previous methods, Nam and Han \cite{nam2016learning} try to leverage a shallow CNN, which is pretrained on a tracking-specific dataset. The pretrained CNN is composed of a shared and sequence-specific part. The shared section extracts the generic feature representation of each sequence. While sequence-specific section (contains multiple branches of sequence-specific layers) differentiates the target from the background in each sequence.

\subsection{Generative CNN-based Trackers}
Generative CNN-based trackers aim to find the most resembling candidate to the target via a generic matching function. One of the outstanding networks in generative approaches undoubtedly pertains to the Siamese networks, which are increasingly being used in visual object tracking. Bertinetto et al.\cite{bertinetto2016fully} adopt a similarity learning framework with no update strategy to train a fully-convolutional Siamese network and find the target within a large area. Two branches of the Siamese network is linked together via a cross-correlation layer. After training the network, a scalar-valued score map specifies the location of the target. In \cite{chen2017once}, the target image and a large search area are transmitted to the inputs of a Y-shaped network and, consequently, the network outputs a binary response map indicating the location of the target. Unlike the network architecture in \cite{bertinetto2016fully}, Chen et al.\cite{chen2017once} concatenate the last three fully connected (\textit{fc}) layers to capture spatiotemporal features \cite{pflugfelder2017siamese}.
Tao et al.\cite{tao2016siamese} prefer to supplant the concatenation layers with a normalization layer before the loss layer in comparison with the other approaches  \cite{bertinetto2016fully,chen2017once}. By this way, authors in \cite{tao2016siamese} attempt to sustain the direction of the feature, while embedding them on a unit sphere. Despite the successes of the aforementioned trackers, they can not satisfactorily cope with large object appearance variations due to the parameters sharing in the Siamese network \cite{pflugfelder2017siamese}. In this paper, a  weighting strategy is incorporated into the Siamese network to make the tracker more versatile in dealing with large object appearance variations. It is worth noting that our proposed method bears some similarity to a recent paper \cite{zhang2018visual} as they incorporate pretrained Siamese network into sequence-specific updating. However, our approach differs from \cite{zhang2018visual} in several respects.
First, Our proposed Siamese network is based on the SINT \cite{tao2016siamese} architecture that integrates hierarchical features in order to consider both semantic and spatial details. In addition, to efficiently emphasize the fine-to-coarse features quality rather than features scale, the normalization layer is used in the SINT architecture to make features restricted to a unit sphere. While the authors of \cite{zhang2018visual} do not capture fine-to-coarse spatial features.
Second, despite fine-tuning the generic Siamese network in \cite{zhang2018visual} to adapt appearance variations, the candidates are solely evaluated by their distances from the ground truth. While, we utilize an adaptable buffer, which is updated by the best candidates over time to account target appearance changes during tracking process. Third, \cite{zhang2018visual} incorporate sequence-specific information into Siamese tracker with fine-tuning the last three layers of the network and by this way they do not increase computational complexity at the cost of losing abstract representation of Siamese network. Compared with \cite{zhang2018visual}, we use a pretrained network WCNN fine-tuned over time for weighting candidates according to their location-specific information.
This network is constituted by two $1\times1$ convolutional layers to reduce the trainable parameters and also preclude network over-fitting.

\section{The proposed Tracker}
\begin{figure}
  \centering
  \includegraphics[scale=0.87]{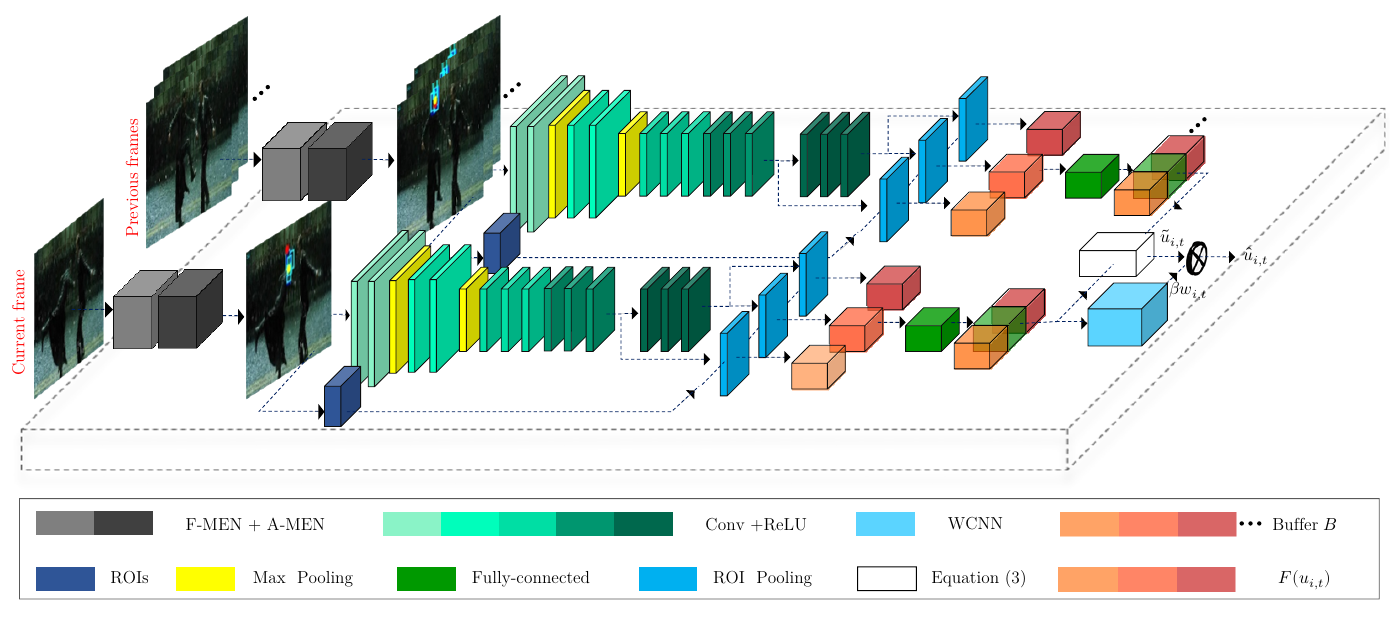}
  \caption{Online tracking framework at time step $t$. This network consists of three components: 1) MEN, which enhances the sampling procedure by providing a further cue (red point) for sampling candidates in addition to the previous target position (yellow point); 2) Siamese network constructed from two identical branches to measure the similarity of its inputs; 3) WCNN, which down-weights distracting candidates using sequence-specific information. The proposed tracking procedure at time step $t (>1)$ is included in Algorithm \ref{tracking_alg}.}\label{fig.Wnet}
\end{figure}

In this section, our proposed method is thoroughly explained. First, MEN is described in depth and then the contribution of baseline Siamese network and its details are discussed. Lastly, the strategies of inference and updating are also presented. The whole framework of our proposed method is illustrated in Figure \ref{fig.Wnet}.

\subsection{Motion Estimation Network}
Motion estimation network aims to enhance the performance of the motion model in complex and fast target movements. A conventional approach used in motion model tries to propagate new candidates according to the previous estimated target location using Gaussian distribution. In this approach, it is assumed that the distance of the target from the ground truth does not exceed a predefined threshold. Hence, when a tracker drifts away in complex movements or challenging scenarios, there is not much chance to redetect the target in the following frames. In order to equip our tracker with the redetection scheme, MEN (inspired by \cite{nam2016learning,yang2017deep}) is utilized. As shown in Figure \ref{fig.Wnet}, this network consists of an adaptive and a fixed component, named \textit{F-MEN} and \textit{A-MEN}. The F-MEN pretrained by VGG-M \cite{chatfield2014return} captures the spatial information of an area at the center of the previous target location.
This area is resized to $107\times107\times3$ before entering into the network. The F-MEN is made up of a pretrained convolutional layer which is followed by a Rectified Linear Unit (ReLU) and a Local Response Normalization (LRN). Compared with the frozen F-MEN, the A-MEN tries to leverage the sequence-specific details. Namely, it attempts to adaptively deal with the target appearance variations while benefiting from both the temporal and long-term information of the target in each sequence simultaneously. To this end, it needs to be updated over time. A-MEN consists of two $1\times1$ convolutional layers trained in the first frame. To train A-MEN, the score map (with size of $51\times51\times2$) is fed into a softmax loss layer, which discriminates positive candidates from negative ones. As \cite{bertinetto2016fully,yang2017deep}, the positive and negative candidates in the score map are determined by:

\begin{equation}\label{eq.1}
y[n]=\begin{cases}
    \;2,& \text{if $\|o-o_g\|\leq R$}\\
    \;1,& \text{otherwise},
\end{cases}
\end{equation}
where $o_g$ and $R$ stand for the center of the previous predicted location and a predefined radius that separates positive candidates from negative ones. On the score map, positive candidates are located within the radius R of the $o_g$. To remove class imbalance, the losses are also weighted.\par
\begin{figure}[!ht]
  \centering
  \includegraphics[scale=0.8]{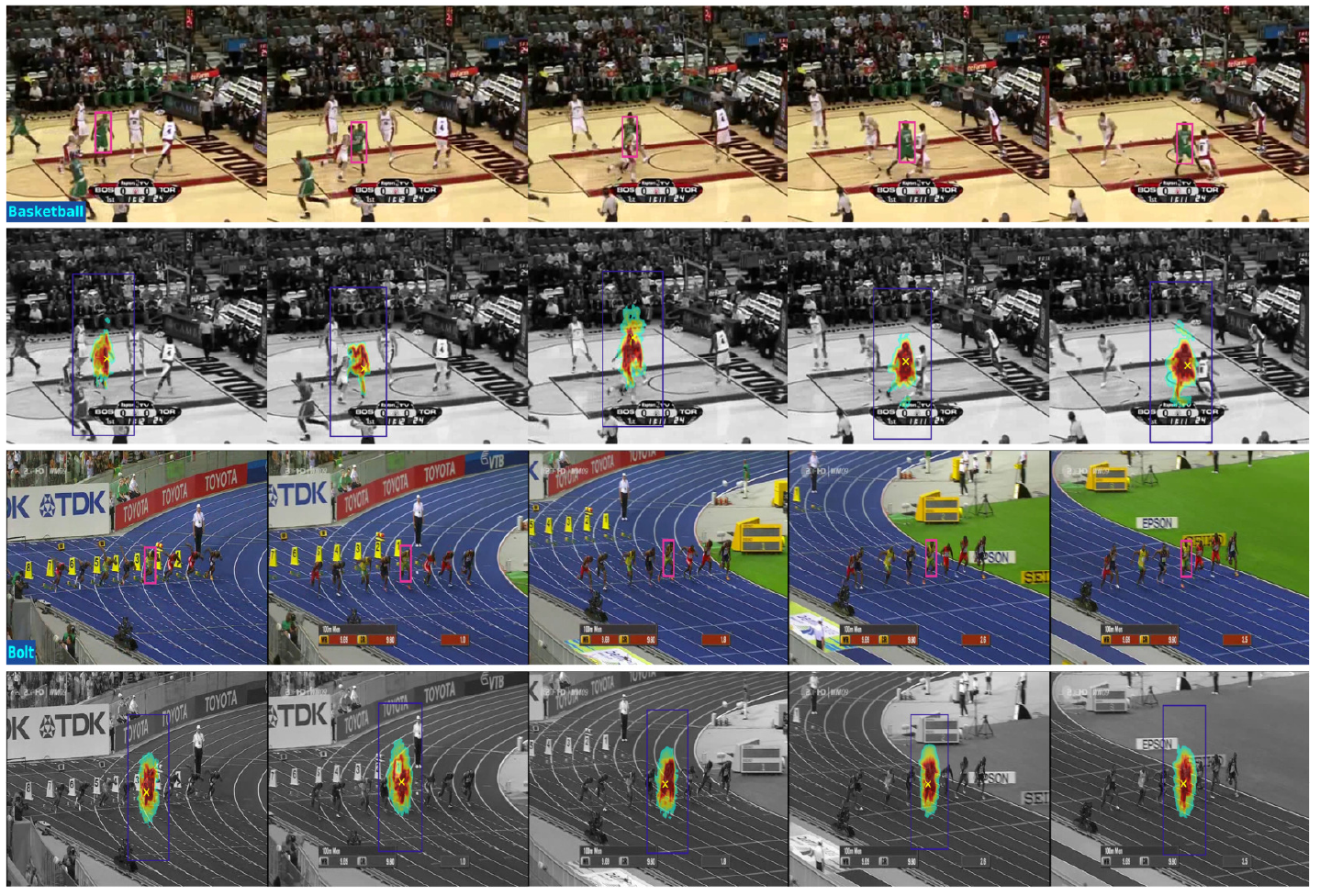}
  \caption{In the even rows, the heat maps are obtained by mapping score maps of the MEN to the input frames (the odd rows). The ground truth bounding boxes and the search regions are represented by pink and blue colors in odd and even rows, respectively. The blue boxes are about four times greater than the size of the corresponding pink boxes. The yellow cross corresponds to the maximum value in the score map. Best viewed in color and magnification. }\label{fig.heatmap}
\end{figure}

After training the A-MEN, the score map of the object of interest is back-propagated to the input image in order to find the most probable region. Figure \ref{fig.heatmap} depicts the results of the back-projection for two sequences. Consequently, new candidates can be generated employing Gaussian distribution at the center of two points: 1) The location in the original input image corresponding to the maximum point in the score map of the MEN. This location is shown with a yellow cross in Figure \ref{fig.heatmap}; 2) The center of the ground truth bounding box (from the second frame on, this location is obtained employing the Siamese network). Equipped with these two plausible locations, we can presume that candidates are subject to a Gaussian distribution around the locations.
Let $u_t=(l_x,l_y,s)$ denote the state variable, where $l_x,l_y$ and $s$ indicate the center coordinate and the scale of the target bounding box, respectively.
Then, the target candidates are drawn using Gaussian distribution as  $\mathcal{N}(u_t,\hat{u}_t^m,\Sigma)$ around the first location and  $\mathcal{N}({u}_t,{\hat{u}}_{t-1}^f,\Sigma)$ around the second location, where $\Sigma$ denotes a diagonal covariance matrix. Afterwards, the total generated candidates are fed into the Siamese network to determine which candidate resembles the target more closely.

\subsection{Baseline Siamese Network}
Siamese network has recently drawn much attention as an approach of similarity learning in visual object tracking.
The convolutional architecture of our proposed method for similarity learning is constructed based on the Siamese network in the SINT \cite{tao2016siamese} architecture, which contains two identical convolutional branches inherited from VGGNet \cite{DBLP:journals/corr/SimonyanZ14a}. This network aims to detect the most similar candidate to the target. Although training a CNN with multiple pooling operations makes the network more robust in dealing with object deformation and also helps to preclude over-fitting, pooling operation decreases the spatial resolution details. For instance, employing five pooling layers in the VGGNet would downsample the spatial information by a factor of $32$. Therefore, the first two pooling layers in the VGGNet are intentionally retained and the other pooling layers, as well as all \textit{fc} layers, are omitted. The architecture parameters of our baseline Siamese network is summarized in Table \ref{table.1}\footnote{Merely one branch of the network is reported.}. To benefit from multiple levels of abstract representation, three ROI (region-of-interest) pooling layers are applied to different layers. These ROI pooling layers transform features of multiple candidates with non-uniform sizes into the fixed-size feature maps simultaneously \cite{girshick2015fast}. Due to the usage of the same input feature map for multiple candidates, ROI pooling layers speed up the performance of the network. Integrating these ROI pooling layers cannot be directly performed via a single concatenation operation, since their scales may differ considerably. Hence, a $\ell_2$ normalization layer is employed after roipool1, roipool3, fc and concatenation layers to make the outputs of different layers
limited to a unit sphere. Eventually, normalized concatenation layers are applied to a loss function, called margin contrastive loss. The underlying aim of this loss function is to make similar inputs embedded close together and to push dissimilar inputs apart \cite{hadsell2006dimensionality}. This loss function is formulated as follows \cite{chopra2005learning}:
\begin{equation}\label{eq.2}
 \ell(u_i,u_j,t_{ij})=\frac{1}{2}y {D}^2 + \frac{1}{2}(1-c_{ij})\max{(0,\tau-{D}^2)}
\end{equation}
Where $c_{ij}$ denotes the similarity of two inputs (i.e $u_i$ and $u_j$) using a binary label; $D=\|F(u_i)-F(u_j)\|_2$  calculates the Euclidean distance between the two normalized outputs of the network; and $\tau$ indicates the minimum margin to separate two distinct inputs.\par
Once the Siamese network has been trained, the most likely candidates can be located using an inference scheme. However, because of evaluating candidates via a fixed Siamese network, it tends to drift in the cases of large appearance changes or coexisting of confusing objects. To tackle these challenges, we pursue two procedures: 1) Unlike previous Siamese-based trackers \cite{bertinetto2016fully,tao2016siamese} relying solely on the initial ground truth to find the target, in this work the best previous candidates are stored in a buffer and the new candidates are then assessed with the stored elements. To impede entering the occluded candidates into the buffer, buffer updating is restricted to a predefined condition; 2) Recently, much attention has been given to the use of sequence-specific information for fine-tuning a pretrained CNN \cite{nam2016learning,yang2017deep,wang2015visual,hong2015online,zhang2018visual,fan2017sanet}.
\begin{table}[!t]
\centering
\tabcolsep 4pt
\caption[Caption for LOF]{The architecture parameters of one branch of the Siamese network. As VGGNet, all convolutional layers are chased by a ReLU layer. Furthermore, roipool1, roipool3, fc and concat layers are also chased by a $\ell_2$ normalization layer, where 'concat' denotes the concatenation layer. The kernel sizes are arranged as: (filter height, filter width, number of filters in the bank).}
\medskip\small
\resizebox{\textwidth}{!}{%
\begin{tabular}{cccccc}
\hline
  \hline
  \addlinespace
  \textbf{Layer} & \textbf{Kernel size/Subdivisions}  & \textbf{Stride/Spatial scale} & \textbf{Input} & \textbf{Output} & \textbf{Output size} \\
  \addlinespace
  \hline
  \addlinespace
  image & - & - & - & - & $512\times512\times3$  \\
  \addlinespace[1mm]
  conv1 (2 sublayers) & $3\times3\times64$  & 1 & image & conv1 & $512\times512\times64$ \\
  \addlinespace[1mm]
  maxpool1 & $2\times2$  & 2 & conv1 & maxpool1 & $256\times256\times64$ \\
  \addlinespace[1mm]
  conv2 (2 sublayers) & $3\times3\times128$  & 1 & maxpool1 & conv2 & $256\times256\times128$\\
 \addlinespace[1mm]
  maxpool2 & $2\times2$  & 2 & conv2 & maxpool2 & $128\times128\times128$ \\
  \addlinespace[1mm]
  conv3 (3 sublayers) & $3\times3\times256$  & 1 & maxpool2 & conv3 & $128\times128\times256$\\
  \addlinespace[1mm]
  conv4 (3 sublayers) & $3\times3\times512$  & 1 & conv3 & conv4 & $128\times128\times512$ \\
  \addlinespace[1mm]
  conv5 (3 sublayers) & $3\times3\times512$  & 1 & conv4 & conv5 & $128\times128\times512$ \\
  \addlinespace[1mm]
  roipool1\footnotemark\setcounter{auxFootnote}{\value{footnote}} & $7\times7$  & 0.25  & conv4 & roipool1 & $1\times1\times25088$\\
  \addlinespace[1mm]
  roipool2 & $7\times7$   & 0.25 & conv5 & roipool2 & $7\times7\times512$\\
 \addlinespace[1mm]
  roipool3\footnotemark[\value{footnote}] & $7\times7$   & 0.25 & conv5 & roipool3 & $1\times1\times25088$\\
  \addlinespace[1mm]
  fc & $7\times7\times4096$  & 1  & roipool2 & fc & $1\times1\times4096$\\
  \addlinespace[1mm]
  concat & -   & - & (roipool1,roipool3,fc) & concat & $1\times1\times54272$ \\
  \addlinespace[1mm]
  \hline
  \hline
  \label{table.1}
\end{tabular}}
\end{table}
\footnotetext{roipool1 and roipool3 are flattened.}Integrating online fine-tuning strategy into a pretrained network can make the tracker adaptive to the large appearance changes of the target and thus enhance the tracker robustness.
The trackers using this strategy locate the object of interest via the maximum positive score obtained from the fine-tuned network. Compared with the aforementioned strategy, in this work the WCNN is employed to learn a weighting mechanism by utilizing sequence-specific information. The weights are generated from the scores of the WCNN. This weighting mechanism can be utilized not only for down-weighting distracting candidates of the similar class but also for taking the target appearance variations into consideration. To achieve this goal, the WCNN constructed from two $1\times1$ convolutional layers is trained by a softmax logistic regression in the initial frame. During tracking, the WCNN would only be fine-tuned under peculiar conditions to preclude the tracker from drifting to similar objects. When the conditions are satisfied, the WCNN would be fine-tuned using features of positive and hard negative candidates ($F(u_{i,t})$).\par

\subsubsection{Training details}
Training deep CNNs requires a large-scale dataset to achieve satisfactory results. ALOV \cite{smeulders2014visual} is one of the standard datasets which consists of 350 sequences with  65 different types of object. It also contains 13 different aspects of difficulty \cite{smeulders2014visual}. To unbiasedly train the Siamese network, all overlapped sequences between training and evaluation datasets should be omitted from the training dataset. Since we use the OTB100, OTB50 \cite{wu2015object} and OTB2013 \cite{wu2013online} datasets for evaluating our proposed tracker, 12 overlapped sequences should be eliminated. We cannot directly use standard datasets for training the proposed Siamese network, and hence we need to make some modifications to the standard datasets. Siamese networks require pairs of inputs images for training operation. To satisfy this requirement, the ground truth bounding box is extracted from a frame of a sequence for the first input. Then, multiple candidates are extracted from other random frames of that sequence for the second inputs. In summary, all candidates in the second inputs are compared with the ground truth in the first input. These candidates are supposed to be positive when their intersection-over-union (IoU) overlap ratios (with the ground truth box in the first input) exceed 0.7. On the other hand, they are considered to be negative if their overlaps do not exceed 0.5.\par
The proposed Siamese network is initialized with the VGG-16 \cite{DBLP:journals/corr/SimonyanZ14a}.
Fine-tuning the proposed Siamese network with the same learning rate for all layers may not lead to the optimal convergence \cite{singh2015layer}. Thus, layer-specific learning rate is utilized. Since initial layers of VGGNet extract low-level features that can be used either for image recognition or visual tracking in the same way, the first nine convolutional layers are frozen. The layer-specific learning rate for all other layers are set to 0.01 except for the fc layer, in which the learning rate is set to 1. All training details of the Siamese network are kept the same as in \cite{tao2016siamese}.

\subsection{Inference and Updating Scenario}
\paragraph{Inference}Once the Siamese network has been trained, it can be regarded as a matching function to locate the most similar objects to the predefined template. In this work, the most probable candidates are stored in the adaptable buffer $B$, which is used as a template in the matching function. This buffer is updated only when the final score ($ \hat{u}_{t}^{f}$) reaches a predetermined threshold (see Algorithm \ref{tracking_alg}). Thereby, occluded candidates are effectively filtered out. To carry out this strategy, the inference strategy in \cite{jiang2018siamese} is modified to take advantage of the adaptable buffer as follows:
\begin{equation}\label{eq.3}
  \tilde{u}_{i,t}=(\eta M(u_g,{u}_{i,t}) +( \frac{1-\eta}{N})\sum_{i=1}^{N} M(b_{i},{u}_{i,t}))
\end{equation}
where the matching function $M$ is formulated as $M(p,q)=F(p)^T F(q)$; $b_{i}$ indicates the $i$-th element of the buffer $B=\{b_1,...,b_N\}$; $\eta$ and $u_g$ are a predetermined threshold and the ground truth in the initial frame; $u_{i,t}$ denotes  $i$-th sampled candidate at time $t$.
Once the sampled candidates have forwarded through the Siamese network, the WCNN can be fine-tuned using the output of the Siamese network while leveraging sequence-specific information. To be more specific, we take the output of the Siamese network $F(u_{i,t})$ to fine-tune and train the WCNN.
Ultimately, to down-weight background distracting candidates, the two scores are integrated as:
\begin{equation}\label{eq.4}
 \hat{u}_{i,t}= \exp({ \beta{{w}}}_{i,t})\odot \tilde{u}_{i,t}
\end{equation}
where $\tilde{u}_{i,t}$ and ${w}_{i,t}$ indicate the $i$-th output scores of the Siamese and WCNN; $\hat{u}_{i,t}$ denotes the final state of $i$-th sampled candidate in frame $t$; $\beta$ is a predetermined threshold. Finally, the final state of the target $\hat{u}_{t}^{f}$ is predicted by averaging the states of top five candidates ($\hat{u}_{i,t}$).

\begin{spacing}{1}
\begin{algorithm}[!t]
  \begin{algorithmic}[1]
    \Input{ Pretrained F-MEN, Pretrained Siamese network, The first ground truth bounding box $u_g$.}
    \Output{Targat location  ${\hat{u}}_{t}^f$,
    \newline Updated: Buffer $B$, A-MEN, WCNN}
    \State Feed $u_g$ to the Siamese network and construct $B$ using $F(u_g)$
    \State Randomly initialize the weights of WCNN and A-MEN.
    \State \multiline{Generate candidates around the first target position $u_g$ and pass them through the F-MEN and Siamese network.}
    \State \multiline{Fine-tune WCNN using $\{F(u_{i,1})\}_{i=1}^{{n}_{1}^{+}}$ and $\{F(u_{i,1})\}_{i=1}^{{n}_{1}^{-}}$.}
    \State Fine-tune A-MEN using the outputs of the F-MEN.
    \State \multiline{  $\mathcal{S}_{short}\leftarrow{1}$ , $\mathcal{S}_{long}\leftarrow{1}$ and $\hat{u}_{1}^f\leftarrow{u_g}$.}
    \For {$t = 2,3,...$  }
        \State \multiline{Apply search window centered at $\hat{u}_{t-1}^f$ to the MEN and output $\hat{u}_t^m$.}
        \State \multiline{Extract sample candidates around $\hat{u}_t^m$ and $\hat{u}_{t-1}^f$.}
        \State \multiline{Compute $\{F(u_{i,t})\}_{i=1}^{256}$ and also $ \tilde{u}_{i,t}$ using Equation (\ref{eq.3}).}
        \State \multiline{Pass $\{F(u_{i,t})\}_{i=1}^{256}$ to WCNN and output $\{{w}_{i,t}\} _{i=1}^{256}$.}
        \State \multiline {Compute $\hat{u}_{i,t}$ and $\hat{u}_{t}^{f}$ using Equation(\ref{eq.4}).}
        \If {$\hat{u}_{t}^f > 1.6 $ or $t<4$}
            \State Update buffer $B$ with the best $\{F(u_{i,t})\}$ corresponding to the $\hat{u}_{t}^{f}.$
            \State \multiline{Generate candidates around the $\hat{u}_{t}^{f}$ and pass them through the F-MEN and Siamese network.}
            \State $\mathcal{S}_{short}\leftarrow\mathcal{S}_{short}\cup\{t\}$, $\mathcal{S}_{long}\leftarrow\mathcal{S}_{long}\cup\{t\}.$

            \If {$|\mathcal{S}_{long}|>{\tau}_{long}$} {$\mathcal{S}_{long}\leftarrow\mathcal{S}_{long}\backslash\{ \min\nolimits_{\forall s \in {\mathcal{S}_{long}}} s \}.$}
             \EndIf

            \If {$|\mathcal{S}_{short}|>{\tau}_{short}$} {$\mathcal{S}_{short}\leftarrow\mathcal{S}_{short}\backslash\{ \min\nolimits_{\forall s \in {\mathcal{S}_{short}}} s \}.$}
            \EndIf
        \EndIf
        \If {$\hat{u}_{t}^f < 1.6 $ }
            \State \multiline{Update WCNN and A-MEN using ${n}_{t}^{+} {_{\forall s \in {\mathcal{S}_{short}}}}$ and ${n}_{t}^{-} {_{\forall s \in {\mathcal{S}_{short}}}}$}
        \ElsIf {$mod(t,{\tau}_{int})=0$}
            \State \multiline{Update WCNN and A-MEN using ${n}_{t}^{+} {_{\forall s \in {\mathcal{S}_{long}}}}$ and ${n}_{t}^{-} {_{\forall s \in {\mathcal{S}_{short}}}}$}
        \EndIf
    \EndFor
  \end{algorithmic}
  \caption{Online tracking algorithm}
  \label{tracking_alg}
\end{algorithm}
\end{spacing}

\paragraph{Updating}During tracking process, the object of interest undergoes a wide range of changes in illumination, pose, scale, orientation and shape. To overcome these challenges, the tracker should be updated over time. Hence, our tracker is equipped with two adaptable memories which are updated in short- and long-term strategies. These strategies can be dated back to the MDNet tracker \cite{nam2016learning}.
In long-term strategy, updating is carried out every ${\tau}_{int}$ frames exploiting positive instances aggregated for ${\tau}_{long}$ time duration in the frame set $\mathcal{S}_{long}$. On the other hand, short-term updating is executed when $ \hat{u}_{t}^{f}$ does not reach a predetermined threshold. The positive instances in short-term updating are aggregated for ${\tau}_{short}$ time duration in the frame set $\mathcal{S}_{short}$. Note that in both strategies we employ negative instances aggregated for ${\tau}_{short}$ time duration.\par
The aforementioned memories play a key role to update two different sections of our proposed tracker: (1) The A-MEN needs to be fine-tuned to properly model the motion of the target. Thereby, the first memory is constructed by the features of positive candidates in $\mathcal{S}_{long}$ and $\mathcal{S}_{short}$, which are calculated by the F-MEN; (2) To capture target appearance changes during tracking, the WCNN should be updated over time. To this end, $F(u_{i,t})$ is obtained for all positive and negative candidates in $\mathcal{S}_{long}$ and $\mathcal{S}_{short}$ to build the second memory.

\section{Experiments}
In this section, we first discuss experimental settings and implementation details. Then, we perform quantitative and qualitative experiments to assess our proposed tracker.
\subsection{Experimental Settings}
To demonstrate the robustness and adaptiveness of our proposed tracker against state-of-art trackers, we conduct extensive experiments on three challenging datasets including OTB100, OTB50 \cite{wu2015object} and OTB2013 \cite{wu2013online}. OTB50 and OTB2013 are subsets of OTB100 dataset. Note that OTB50 encompasses more challenging sequences than OTB2013. The videos in these datasets are labeled with 11 attributes such as motion blur, occlusion, deformation, fast motion, background clutter and so forth. We adopt two metrics in OTB toolkit \cite{wu2015object}, i.e., distance precision (DP) and overlap success (OS), to draw precision and success plots and evaluate our tracker against the state-of-art competitors.
The precision plot illustrates the ratio of frames where the distance of central pixels between the predicted and the ground truth bounding boxes does not reach to a predefined threshold. On the other hand, the success plot gauges the percentage of frames that their overlap criteria exceed a predefined threshold. By changing the threshold between 0 to 50 (0 to 1), the precision (success) plot is obtained. To rank trackers based on the distance precision and overlap success rates, the threshold of 20 pixels and \textit{area under curve} (AUC) are utilized, respectively. We evaluate our proposed tracker against the state-of-art trackers including VITAL \cite{song2018vital}, MDnet \cite{nam2016learning}, ADnet \cite{yoo2017action}, HCFTs \cite{ma2017robust}, SRDCF \cite{danelljan2015learning}, CREST \cite{song2017crest}, SiamFC \cite{bertinetto2016fully}, ACFN \cite{choi2017attentional}, CFNet \cite{valmadre2017end}, Staple \cite{bertinetto2016staple}, LCT \cite{ma2015long}.
\subsection{Implementation Details}
To distinguish the positive candidates from the negative ones, the threshold $R$ in Equation (\ref{eq.1}) is set to 12. The A-MEN is trained (fine-tuned) for 30 (10) epochs using stochastic gradient descent (SGD) with a global learning rate of 0.001, a momentum of 0.9, and a batch size of 8. The layer-specific learning rate for the first and second convolutional layers are set to 3 and 30, respectively. Overfitting issue is reduced using a weight decay of 0.0005. To preclude boundary discontinuities, the features calculated from the F-MEN are multiplied by a cosine window before inputting them into the A-MEN. The F-MEN remains fixed during the tracking process.
WCNN follows the same training settings as A-MEN, except that the learning rate and momentum are set to 0.15 and 0.005, respectively. Moreover, each mini-batch in training the WCNN contains 32 positives and 96 hard negatives examples. In each frame, 250 candidates are sampled from the state variable ${x_t}$ using Gaussian distribution around the expected locations centered at $ \hat{u}_t^{f}$  and  $\hat{u}_t^m$.
The diagonal covariance matrix $\Sigma$  is formulated as $\textmd{diag}(0.09v^2,0.09v^2,02.5)$, where $v$ represents the mean of bounding box size. To
calculate the scale of new candidates in each frame, the ground truth scale is multiplied by a factor of $1.1$ over time.

\paragraph{Candidate Generation} For training the WCNN, candidates are supposed to be positive when their IoU overlap ratios with the ground truth exceed 0.7. The numbers of positive and negative candidates (${n}_{t}^{+}$ and ${n}_{t}^{-}$) for training the WCNN are set to 50 and 200. Likewise, they are considered to be negative if their IoU overlap ratios do not exceed 0.5. In fine-tuning phase, candidates are deemed to be negative if their IoU overlap ratios do not reach 0.3. Positive candidates follow the same assumption as before and the number of positive and negative candidates (${n}_{1}^{+}$ and ${n}_{1}^{-}$) are set to 500 and 5000, respectively. Furthermore, to train and fine-tune the A-MEN, 50 and 5 positive candidates are accidentally chosen from ${n}_{t}^{+}$ and ${n}_{1}^{+}$ . The parameters used in long and short-term strategies, i.e., ${\tau}_{long}$, ${\tau}_{short}$ and ${\tau}_{int}$ are set to 100, 20 and 10, respectively. $\tau$ in Equation (\ref{eq.2}) , $N$ and $\eta$ in Equation (\ref{eq.3}) and at last $\beta$ in Equation (\ref{eq.4}) are set to 1, 35, 0.7 and 0.2, respectively.

\subsection{Quantitative Comparison}
\begin{figure}[!t]
  \centering
  \includegraphics[scale=1]{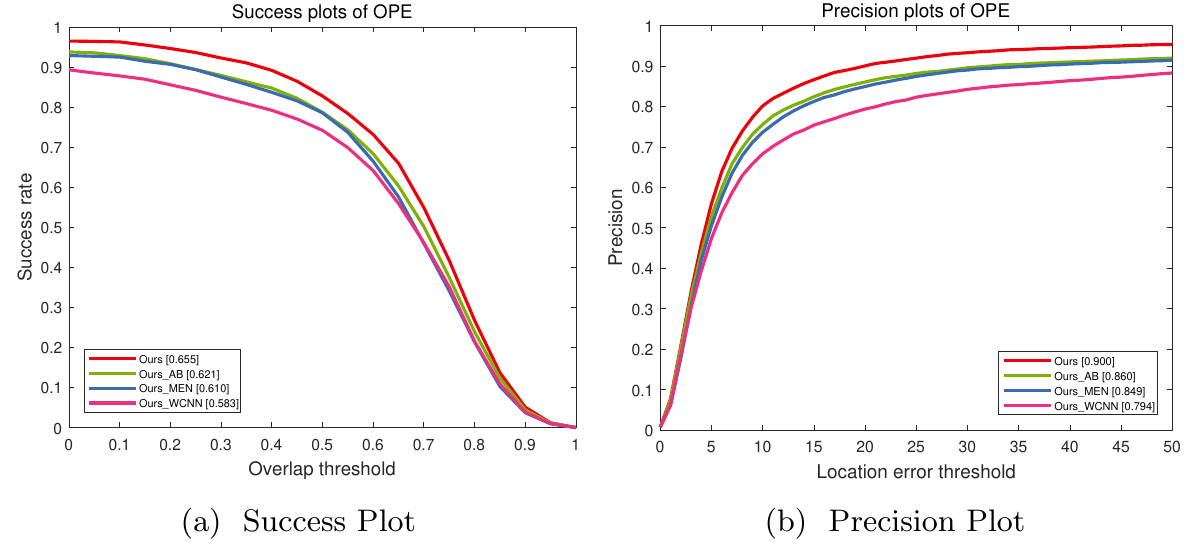}
  \caption{Self-comparison results in terms of success (a) and precision (b) plots on the subset of OTB100 dataset, including the 35 most challenging sequences. }\label{fig.self_comparison}
\end{figure}

\paragraph{Self-comparison}
To investigate the contribution of MEN, WCNN and the adaptable buffer B, we evaluate the degraded versions of our tracker by removing them separately. In Figure \ref{fig.self_comparison}, we denote these three versions as Ours-MEN, Ours-WCNN and Ours-AB. As depicted in Figure \ref{fig.self_comparison}, each component in the proposed method plays an important role to improve the tracking accuracy. Particularly, the WCNN has the most influence on the tracking results.

\begin{figure}[!t]
  \centering
  \includegraphics[scale=0.9]{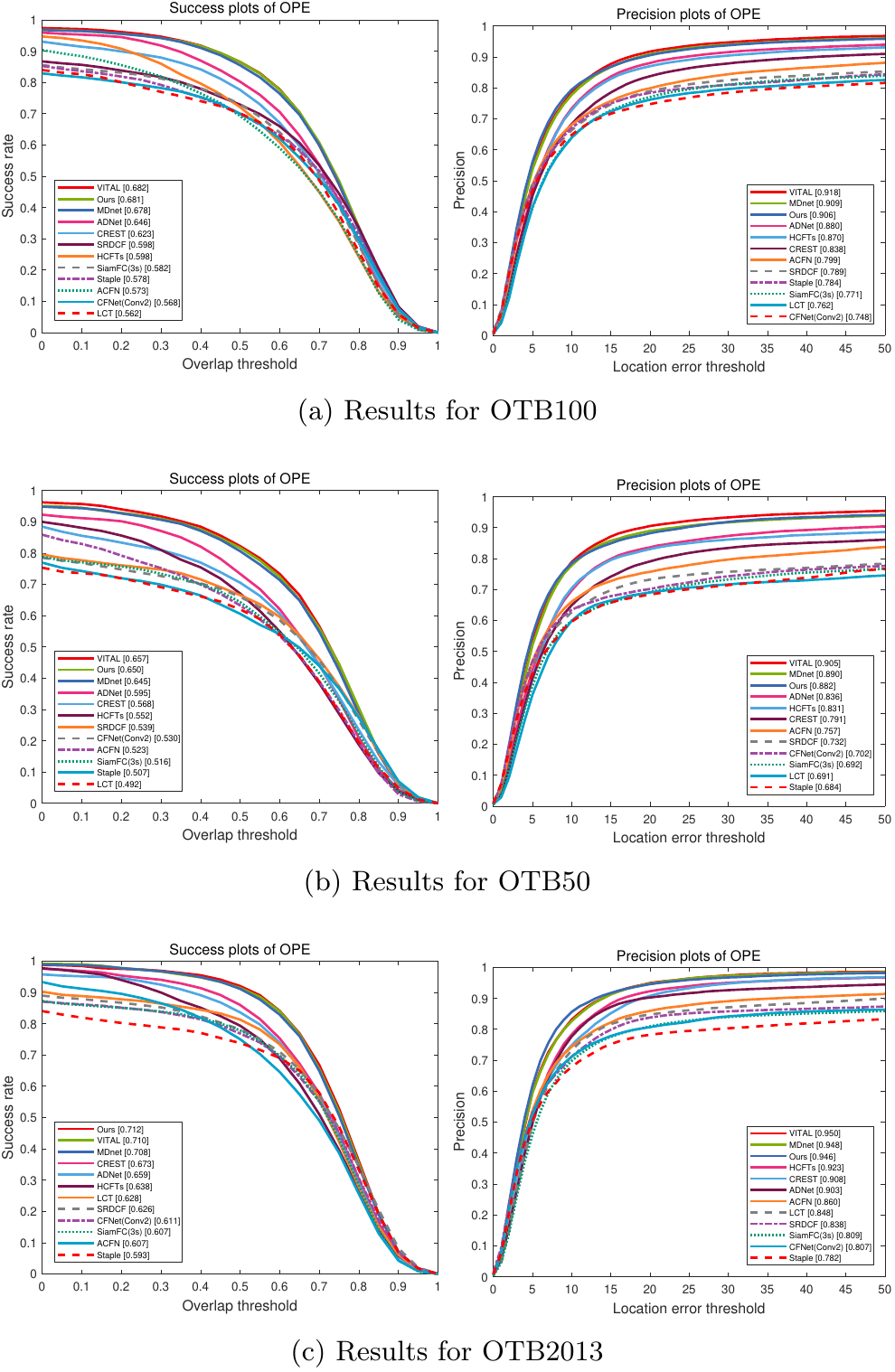}
     \caption{Overall results of OPE according to the DP and OS rates on three datasets, including OTB100, OTB50 and OTB2013. The AUC scores and the DP rates (at the
threshold of 20 pixels) are represented in the legend of subfigures. Only top twelve trackers are shown.}\label{fig.overalReslts}
\end{figure}

\paragraph{Overall Results}
We conduct \textit{one pass evaluation} (OPE) of OTB toolkit to assess our tracker on large-scale OTB benchmark. Figure \ref{fig.overalReslts} illustrates the overall performance of the proposed tracker according to the DP and OS rates on OTB100, OTB50 and OTB2013. As shown in Figure \ref{fig.overalReslts}, the proposed tracker performs satisfactorily in comparison with the others competitors. Figure \ref{fig.overalReslts} demonstrates that our tracker achieves better results in terms of OS criterion in comparison with the DP criterion. For the sake of clarify, in Table \ref{table.2}, the quantitative results of precision and success plots are summarized. It is observed that we obtain a gain of \%0.3 (\%0.5, \%0.4) AUC score over the MDNet on OTB100 (OTB50, OTB2013), respectively. This superiority is also maintained over the VITAL on OTB2013.
In summary, our tracker gets the second best result in terms of overlap success rate and also the third best result in terms of distance precision rate.

\begin{table}[!t]
\centering
\tabcolsep 3pt
\caption[Caption for Quantitative results]{Quantitative results on OTB100, OTB50 and OTB2013. Overlap success (OS) and distance precision (DP) are reported according to the AUC score and  the error threshold of 20 pixels (in Figure \ref{fig.overalReslts}), respectively. Top three results are indicated in \textcolor{red}{red}, \textcolor{blue}{blue} and \textcolor{green}{green} font. }
\medskip\small
\begin{tabular}{LcLcLcL}
\hline
  \hline
  \addlinespace
   & \multicolumn{2}{L}{{OTB100}} & \multicolumn{2}{l}{{OTB50}} & \multicolumn{2}{l}{{OTB2013}}\\
  \addlinespace
  \textbf{Method} & {DP}  & {OS} & {DP} & {OS} & {DP}& {OS} \\
  \addlinespace
  \hline
  \addlinespace
  Ours  & \color{green}90.6 & \color{blue}68.1 & \color{green}88.2 & \color{blue}65 & \color{green}94.6 & \color{red} 71.2 \\
    \addlinespace[1mm]
  VITAL \cite{nam2016learning} & $\color{red} 91.8$  & $\color{red} 68.2$ & $\color{red} 90.5$ & $\color{red} 65.7$ & $\color{red} 95.0$ & $\color{blue}71.0$\\
  \addlinespace[1mm]
  MDNet \cite{nam2016learning} & $\color{blue}90.9$  & $\color{green}67.8$ & $\color{blue}89.0$ & $\color{green}64.5$ & $\color{blue}94.8$ & $\color{green}70.8$\\
  \addlinespace[1mm]
  ADnet \cite{yoo2017action} & $88$ & $64.6$ & $83.6$ & $59.5$ & $90.3$ & $65.9$ \\
  \addlinespace[1mm]
  CREST \cite{song2017crest}& $83.8$  & $62.3$ & $79.1$ & $56.8$ & $90.8$ & $67.3$\\
 \addlinespace[1mm]
  HCFTs \cite{ma2017robust} & $87.0$  & $59.8$ & $83.1$ & $55.2$ & $92.3$ & $63.8$ \\
  \addlinespace[1mm]
  \hline
  \hline
  \label{table.2}
\end{tabular}
\end{table}

\begin{figure}[!t]
  \centering
  \includegraphics[scale=1]{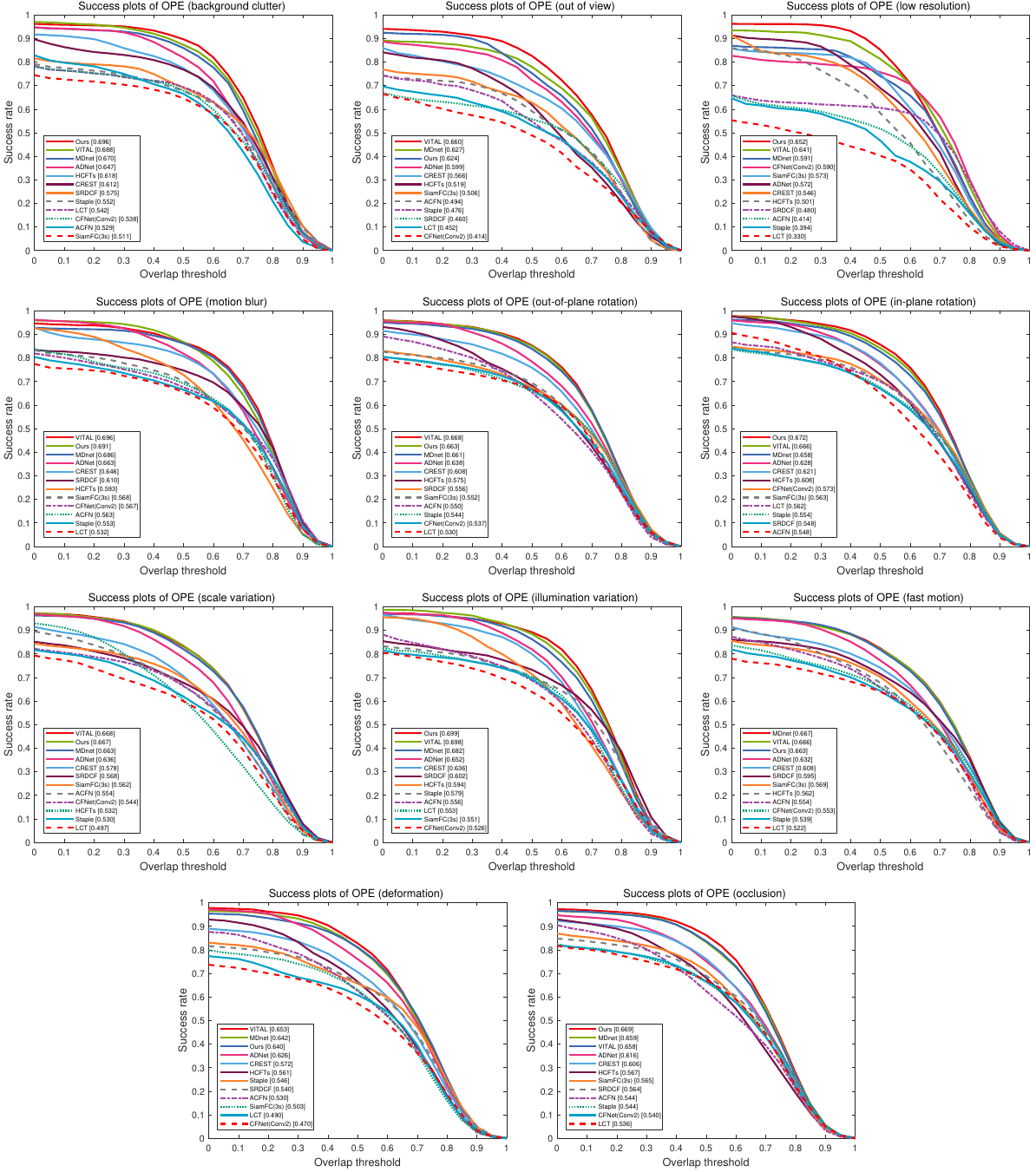}
   \caption{Success plots of OPE on 11 challenges, including fast motion, background clutter, motion blur, deformation, illumination variation, in-plane-rotation, low-resolution, occlusion, out-of-view, out-of-plane-rotation, scale variation.}\label{fig.AttributeReslts_succ}
\end{figure}

\begin{figure}[!t]
  \centering
  \includegraphics[scale=1]{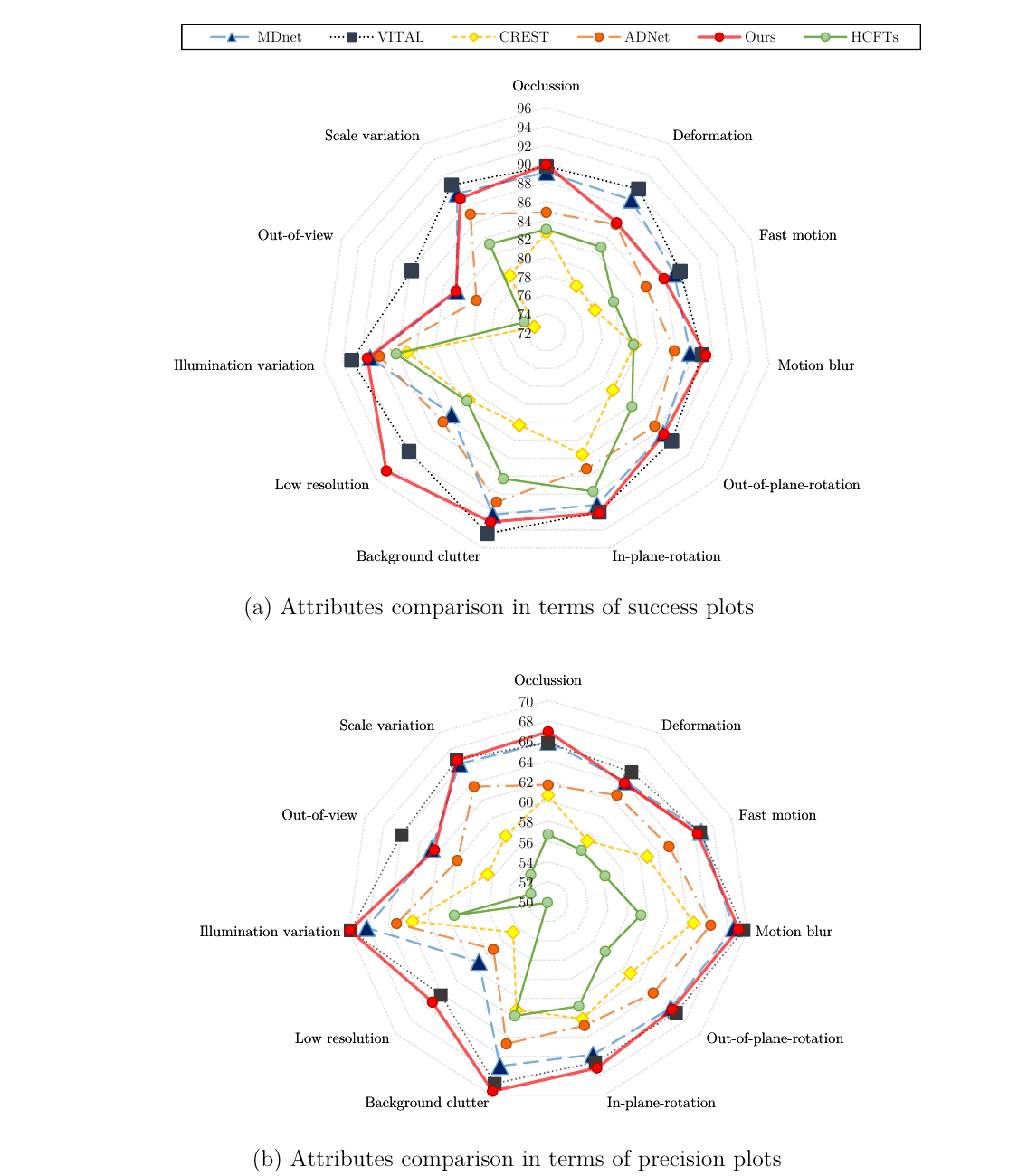}
  \caption{Distance precision rates (a) (at 20 pixels) and AUC score (b) of overlap success rates for the attribute-based results on the OTB100 dataset. }\label{fig.attributeCom}
\end{figure}

\paragraph{Performance Analysis per Attribute}In order to assess the proposed tracker ability in handling different challenges, the success plots for different subsets of the OTB dataset are shown in Figure \ref{fig.AttributeReslts_succ}. Each subset is categorized based on its attributes. Note that a sequence can take multiple tags in the process of tagging. Figure \ref{fig.AttributeReslts_succ} illustrates the success plots of the attribute-based results on the OTB100 dataset. Figure \ref{fig.AttributeReslts_succ} demonstrates that the proposed tracker achieves satisfactory results in tackling all reported challenges, especially in the cases of low resolution and occlusion challenges. For a clearer comparison, the AUC scores in Figure \ref{fig.AttributeReslts_succ} and also the DP rates (at the threshold of 20 pixels on OTB100) are summarized in Figure \ref{fig.attributeCom}. The proposed tracker ranks within top 3 on all 11 challenges.
In terms of both success and precision plots, our tracker ranks the first in low resolution, in-plane rotation and occlusion challenges and obtains \%1.1(\%3.2), \%0.6(\%0.1) and \%1(\%0.2) improvements over the second best trackers, respectively. Due to the usage of hierarchical features, the highest improvements corresponds to the low resolution. Furthermore, Owing to the restrictions imposed on buffer updating, the proposed tracker can better handle occlusion than others. The proposed tracker also ranks the first on the success plots in the background clutter and illumination variation attributes and ranks the second in the out-of-plane rotation, scale variation and motion blur attributes.

\subsection{Qualitative Comparison}

\begin{figure}[!t]
  \centering
  \includegraphics[scale=1]{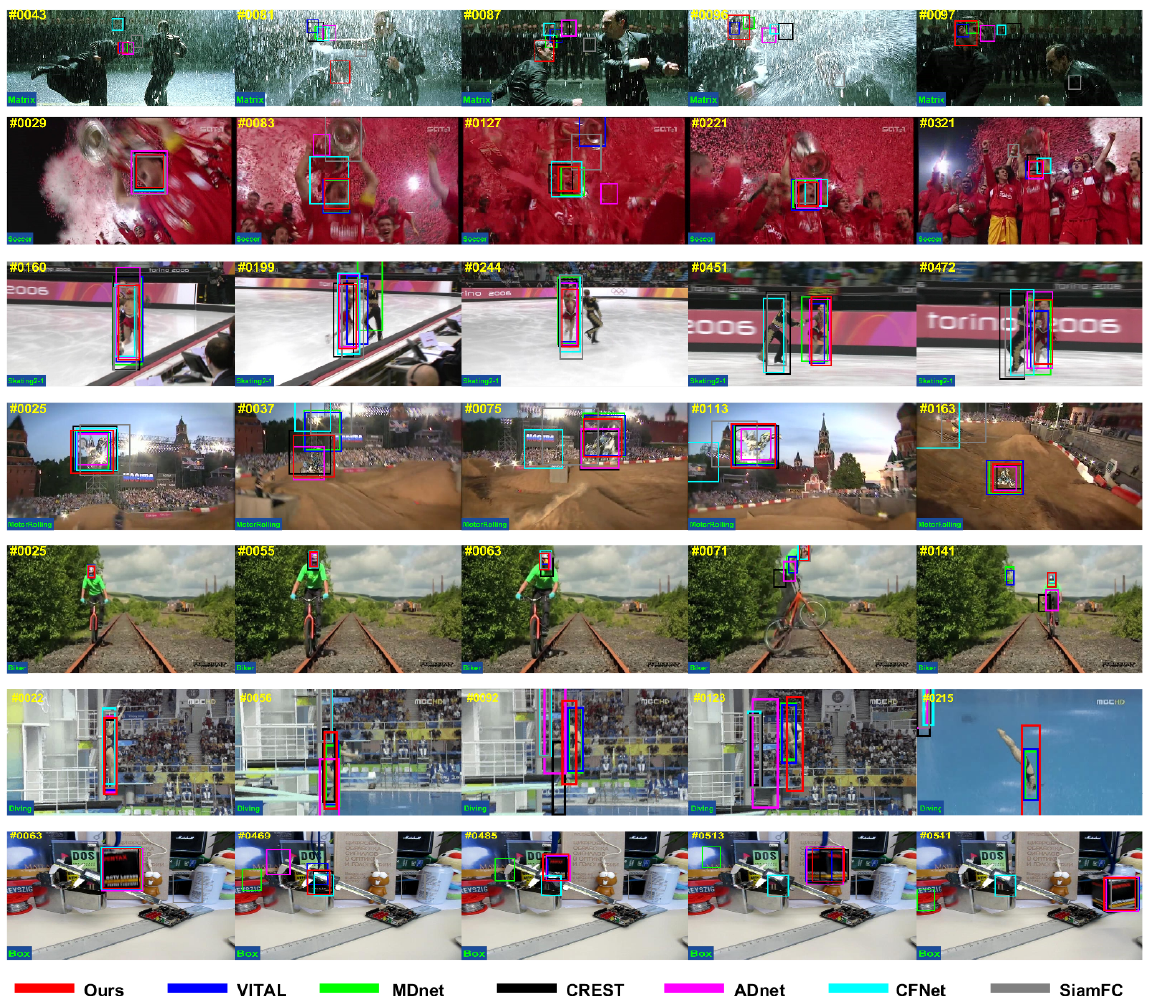}
  \caption{Qualitative performance of the proposed tracker, VITAL \cite{song2018vital}, MDnet \cite{nam2016learning}, ADnet \cite{yoo2017action}, CREST \cite{song2017crest}, SiamFC \cite{bertinetto2016fully}, CFNet \cite{valmadre2017end} on some most challenging sequences from the OTB100 dataset (from top to bottom:  \textit{matrix}, \textit{soccer}, \textit{skating2}, \textit{motorRolling}, \textit{biking}, \textit{diving}, \textit{box})}\label{fig.BoundingBox}
\end{figure}

Figure (\ref{fig.BoundingBox}) illustrates the qualitative performance of the proposed tracker on some most challenging sequences from the OTB100 dataset in comparison with its main competitors, including VITAL \cite{song2018vital}, MDnet \cite{nam2016learning}, ADnet \cite{yoo2017action}, CREST \cite{song2017crest}, SiamFC \cite{bertinetto2016fully}, CFNet \cite{valmadre2017end}.
It is noteworthy that for challenging sequences such as \textit{soccer}, \textit{diving}, \textit{skating2} and \textit{motorRolling}, the trackers exploiting sequence-specific information (i.e., VITAL, MDnet and our proposed tracker) perform more robustly than the other trackers in dealing with occlusion, deformation, scale and illumination variations. However, VITAL and MDnet trackers tend to fail when the target moves fast, such as \textit{biking} sequence. In contrast, our proposed tracker performs more accurately due to the use of MEN. In the box sequence, updating model with occluded objects makes MDnet tracker drift away, whereas our tracker and VITAL perform favorably against long occlusion challenge.
In the \textit{matrix} sequence, where the background is quite cluttered, most approaches drift away. Again, the proposed tracker accurately localizes the target.

\section{Conclusion}
In this paper, we propose a robust tracking approach that addresses the motion and observation models simultaneously. In terms of the motion model, motion estimation network (MEN) is utilized to sample the most probable candidates. Then, to detect the best candidates among all sampled candidates, the Siamese network is trained offline. Due to the target appearance variations during the tracking process, each candidate is evaluated with an adaptable buffer containing the best selected previous candidates. In order to efficiently handle occlusion, buffer updating is limited to a predefined condition. Besides, to make the tracker more robust in dealing with coexisting of similar object and large appearance changes, a weighting CNN (WCNN) is exploited. This WCNN employs sequence-specific information to down-weight distracting candidates. Experimental results verify that our approach performs satisfactorily against the state-of-art trackers.

\bibliography{mybibfile}

\end{document}